\pdfoutput=1
\documentclass[svgnames]{article}

\usepackage[nonatbib, final]{neurips_2024}
\usepackage[numbers, sort, compress]{natbib}

\usepackage{xcolor}
\usepackage{microtype}
\usepackage{hyperref}
\usepackage{url}
\usepackage{booktabs}
\usepackage{graphicx}
\usepackage{color-edits}
\usepackage{booktabs, multirow}
\usepackage{pdflscape}
\usepackage{amsmath, amsfonts, amssymb}
\usepackage{bbm}
\usepackage{subcaption}

\hypersetup{colorlinks=true, citecolor=Navy, linkcolor=blue, urlcolor=blue}

\addauthor[Hanna]{hw}{Teal}
\addauthor[AlexC]{ac}{DarkMagenta}
\addauthor[Luke]{lg}{Blue}
\addauthor[Solon]{sb}{Green}

\title{A Framework for Evaluating LLMs \\ Under Task Indeterminacy}

\author{%
  Luke Guerdan\thanks{Work done as an intern at Microsoft Research.} \\
  Carnegie Mellon University\\
  \texttt{lguerdan@cs.cmu.edu} \\
  \And
  Hanna Wallach \\
  Microsoft Research \\
  \texttt{wallach@microsoft.com} \\
  \AND
  Solon Barocas \\
  Microsoft Research \\
  \texttt{solon@microsoft.com} \\
  \And
  Alexandra Chouldechova \\
  Microsoft Research \\
  \texttt{alexandrac@microsoft.com} \\
}

\begin{document}

\maketitle

\begin{abstract}
Large language model (LLM) evaluations often assume there is a single correct response---a \emph{gold label}---for each item in the evaluation corpus. However, some tasks can be \textit{ambiguous}---i.e., they provide insufficient information to identify a unique interpretation---or \textit{vague}---i.e., they do not clearly indicate where to draw the line when making a determination. Both ambiguity and vagueness can cause \textit{task indeterminacy}---the condition where some items in the evaluation corpus have more than one correct response. In this paper, we develop a framework for evaluating LLMs under task indeterminacy. Our framework disentangles the relationships between  task specification, human ratings, and LLM responses in the LLM evaluation pipeline. Using our framework, we conduct a synthetic experiment showing that evaluations that use the ``gold label'' assumption underestimate the true performance. We also provide a method for estimating an error-adjusted performance interval given partial knowledge about indeterminate items in the evaluation corpus. We conclude by outlining implications of our work for the research community.\looseness=-1 
\end{abstract}

\section{Introduction}

A growing body of research has examined how to evaluate the capabilities and limitations of large language models (LLMs) \citep{hendrycks2020measuring, supernaturalinstructions, wang2018glue, feffer2024red, longpre2024safe, ganguli2022red,shankar2024validates,magooda2023framework}. The majority of evaluations are based on multiple-choice question (MCQ) or question-answering (QA) tasks where it is assumed there is a single correct response---a  \textit{gold label}---for each item in the evaluation corpus.  These gold labels are often obtained by asking human raters to specify the ``correct'' response to each item. While recent work has argued that some tasks, such as stereotype annotation, can be \textit{ambiguous} or \textit{vague}, creating subjectivity that leads to variation in human ratings \citep{bommasani2023holistic, goyal2022your, nadeem2020stereoset, parrish2024diversity, wang2024case}, evaluation designers currently lack practical tools for quantifying the impact of such subjectivity on evaluations of LLMs \citep{fleisig_perspectivist_2024, plank_problem_2022}.

To fill this gap, we develop a framework for evaluating LLMs under task indeterminacy---the condition where some items in the evaluation corpus have more than one correct response, as we further explain below. LLM evaluation under such indeterminacy is challenging because variation in human ratings may reflect either meaningful signal or exogenous error. As a result, our framework begins by disentangling sources of variation in the human rating process used to obtain gold labels. After providing an overview of our framework below, we introduce a method for estimating an error-adjusted \textit{performance interval} given partial knowledge about indeterminate items in the evaluation corpus. \looseness=-1

\section{Related Work}\label{sec:related_work}

A growing body of work spanning HCI \citep{chen_judgment_2023, chang_revolt_2017, bragg_sprout_2018, manam_wingit_2018, davani_d3code_2024, gordon_disagreement_2021, sap_annotators_2022, pavlick_inherent_2019, davani_dealing_2022, binns_like_2017, basile_its_nodate}, NLP \citep{baan_stop_2022,larimore_reconsidering_nodate,prabhakaran_releasing_2021,aroyo_dices_nodate,huang_culturally_2023,basile_its_nodate,binns_like_2017,davani_dealing_2022}, and ML \citep{klie_annotation_2023, resnick_survey_2021, lakkaraju_bayesian_2015, weerasooriya_disagreement_2023, liu2019learning, pavlick2019inherent,peterson2019human,fisch2020efficient} has investigated sources of variation in the human rating process related to task indeterminacy. For example, prior empirical studies have found that individuals from differing cultural or demographic backgrounds often assign different ratings for concepts such as ``toxicity'', ``hate speech'', or ``stereotyping'' \citep{sap_annotators_2022, pavlick_inherent_2019, davani_dealing_2022, binns_like_2017, basile_its_nodate}. These studies have found that aggregating human ratings into a single gold label can cause groups with differing views to be overlooked throughout the model evaluation process \citep{wang2024case}. As a response, the ML and NLP communities have developed modeling frameworks that better account for human rating variation during model training---e.g., by targeting  \textit{soft labels} that represent the distribution of responses assigned during the rating process \citep{liu2019learning, pavlick2019inherent,peterson2019human}. The development of these approaches coincides with a broader ``perspectivist turn'' in the NLP literature \citep{plank_problem_2022, fleisig_perspectivist_2024}, in which human rating variation is viewed as an important signal to be captured throughout both model training and the evaluation process. \looseness=-1

Despite growing awareness of human rating variation and its importance for valid and reliable evaluations, there is a lack of practical frameworks for (1) isolating sources of variation in the human rating process and (2) parameterizing the effect of each source on LLM evaluations \citep{fleisig_perspectivist_2024, plank_problem_2022}. Instead, existing frameworks model human rating variation by training a model to predict soft labels then evaluating the model's predictions against aggregated gold labels \citep{fleisig_perspectivist_2024, plank_problem_2022, uma2020case, geng_label_2016, gao_deep_2017, gao_deep_2017}.  Yet soft label-based training approaches assume that the distribution of human ratings being targeted is a meaningful yardstick of model performance---i.e., that the distribution of ratings provided during data annotation is consistent with the target population of users at deployment time, and that ratings are uncorrupted by exogenous error. Furthermore, while targeting soft labels accounts for human rating variation during training, it does \textit{not} characterize the impact of human rating variation on downstream model evaluations. \looseness=-1

Finally, other frameworks offer more targeted interventions for mitigating the effect of human rating variation on model evaluations. For instance, \citet{gordon2021disagreement} develop a ``disagreement de-convolution'' that enables evaluation designers to correct for noise in human ratings. Yet, at times, raters may provide different responses due to meaningful sources of signal as opposed to exogenous sources of noise. These factors related to task specification are not modeled under this framework. Furthermore, \citet{gordon2022jury} devise an approach that enables evaluation designers to explicitly account for human rating variation connected to demographic factors such as age and gender. Under this framework, a model is trained to predict the rating assigned by each individual in a population (i.e., represented by a combination of demographic factors). At inference time, the system generates a final response by taking a weighted-average of individual-level predictions, where the weighting function is a normative choice made by the evaluation designer. Yet this approach does not examine specific causal mechanisms, such as task ambiguity and vagueness, that might cause raters from different demographic backgrounds to disagree. In contrast, our framework disentangles the influence of several distinct factors---i.e., task specification and rater error---that introduce variation in the human rating process. We develop an approach for quantifying the impact of each of these factors on LLM performance estimates. To our knowledge, our use of causal directed acyclic graphs (DAGs) and performance intervals is novel to the literature studying LLM evaluation under human rating variation.\looseness=-1

 \section{Framework Overview}
 
 Task indeterminacy naturally arises when task instructions are \textit{ambiguous}---i.e., they provide insufficient information to identify a unique interpretation---or \textit{vague}---i.e., they do not clearly indicate where to draw the line when making a determination. Consider, for instance, a harm classification task that instructs raters (either human or AI) to assess whether the statement \textit{``William is such a Cheesehead!''} is ``derogatory toward a person or a group of people.''  This instruction is ambiguous because it permits multiple reasonable interpretations.  Whereas an American rater might view \textit{``Cheesehead''} as an endearing reference to a Green Bay Packers Football fan, and respond \textit{No}, a Dutch rater might connect \textit{``Cheesehead''} to its historical use as a WW2-era pejorative slur, and thus respond \textit{Yes}.   As we explain below, treating just one response as correct when including ambiguous~or~vague questions in an LLM evaluation\footnote{One might argue that ambiguous questions like this simply shouldn't appear in an evaluation corpus.  However, we need to be able to evaluate LLMs on tasks related to safety and harm annotation that are viewed by many as \textit{inherently} subjective \cite{homan2024intersectionality, wilkinson2022many}.   So removing all ambiguous questions is generally not a viable solution. } leads to incorrect performance estimates. \looseness-1

\begin{figure}[!t]
    \centering
    \begin{minipage}[t]{0.6\linewidth}
        \centering
        \includegraphics[trim=6mm 2mm 6mm 2mm, clip,width=\linewidth]{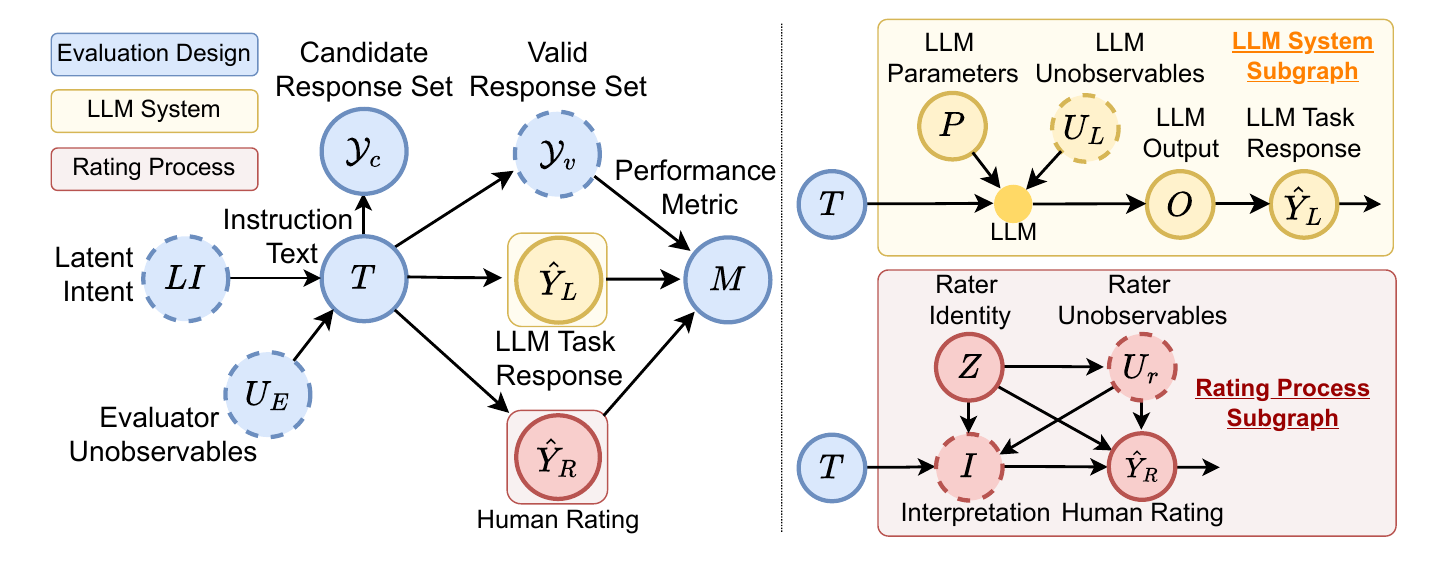}
        \caption{An overview of our causal directed acyclic graph (DAG) for the LLM evaluation pipeline. The right panel expands both the LLM system and human rating process. }
        \label{fig:dag_min}
    \end{minipage}
    \hfill
    \begin{minipage}[t]{0.37\linewidth}
        \centering
        \includegraphics[trim=1mm 1mm 1mm 1mm, clip, width=\linewidth]{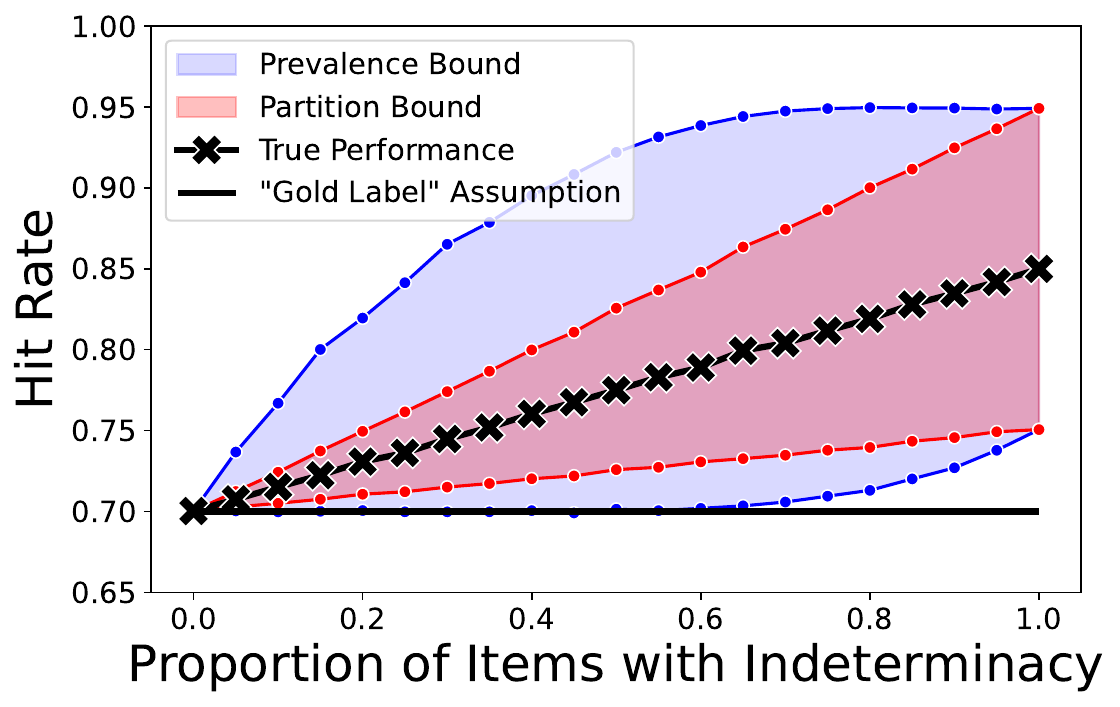}
        \caption{The gold label assumption yields an underestimate of LLM performance under indeterminacy.}
        \label{fig:llm_eval}
    \end{minipage}
\end{figure}

 Our framework, shown in Figure \ref{fig:dag_min}, uses a DAG to describe how task specification, human ratings, and LLM responses affect model performance. The \textit{instruction text}, $T$, is the written task description provided to an LLM or human rater. In the example above, the instruction text would be \textit{``Is the statement `William is such a Cheesehead!' derogatory toward a person or a group of
people? (A) Yes, (B) No.''}  This instruction text is ambiguous. In particular, because human raters observe  \textit{only} the instruction text, two raters might form different interpretations and assign different ``correct'' responses based on their own cultural contexts. This is captured in the \textit{valid response set} (VRS), $\mathcal{Y}_v$, which is the set of all responses that are correct for at least one ``reasonable'' interpretation of the instruction text. In the example above, the VRS would be \{\textit{Yes, No}\} to reflect both reasonable cultural interpretations of the instruction text.  Typical evaluation approaches obtain a single gold label by aggregating ratings from multiple individuals into a single response (Fig \ref{fig:dag_min}; red), which can be viewed as approximating the VRS as a singleton set. However, as we explain below,~this methodology yields a biased estimate of the true performance of a model under task indeterminacy.     \looseness=-1

\section{Evaluating LLMs Under Task Indeterminacy} 

The standard gold label approach to  evaluating LLMs measures performance via the concurrence between the LLM response, $\hat{Y}_L$, and the \text{aggregate} human rating, $\hat{Y}_R$---i.e., $M(\hat{Y}_L, \hat{Y}_R) = \mathbb{P}(\hat{Y}_L = \hat{Y}_R)$. This performance measure only admits one possible response for each item in the evaluation corpus. However, when tasks are indeterminate, an LLM response to an item should be deemed correct if it matches \textit{any} of the responses in that item's VRS.  We therefore define the \textit{true performance} as $M^*(\hat{Y}_L, \mathcal{Y}_v) = \mathbb{P}(\hat{Y}_L \in \mathcal{Y}_v)$. We illustrate the relationship between the true performance and gold label-based evaluations by conducting a synthetic experiment with randomly generated data consistent with our DAG, as shown in Figure~\ref{fig:dag_min}. Figure \ref{fig:llm_eval} shows that evaluations that use the gold label assumption underestimate the true performance.  Furthermore, the magnitude of the evaluation bias increases as the proportion of indeterminate items in the evaluation corpus increases. Intuitively, this is because task indeterminacy introduces additional ways for the LLM's responses to be correct. \looseness=-1

Although in principle we could estimate the true performance by expending effort to obtain (or estimate) the VRS for each item, this approach is only viable for static evaluations, and even then it could be too costly.  We instead propose an alternative approach that uses \textit{partial knowledge} about indeterminate items to bound the true performance. The \textit{prevalence bound} uses an estimate of the proportion of indeterminate items to construct a performance interval. This proportion can be estimated by examining a random sample of items for indeterminacy (e.g., via crowdsourcing techniques \citep{k2019taskmate, chen2023judgment}). The \textit{partition bound} is obtained by splitting the evaluation corpus into two subsets: \textit{determinate} (items with $|\mathrm{VRS}|=1$) and \textit{indeterminate} ( $|\mathrm{VRS}|\ge 1$). One heuristic for partitioning is to sort the items by the level of human-rater (or LLM response) agreement, and then select an agreement threshold below which an item is deemed to be indeterminate.  Figure~\ref{fig:llm_eval} shows the bounds resulting from each approach for varying proportions of indeterminate items.  We see that the partition bound is much narrower than the prevalence bound, because it relies on having additional information about which items are indeterminate. Although existing frameworks provide mechanisms for evaluating ML models under human rating variation arising from vagueness or ambiguity \citep{gordon2021disagreement, gordon2022jury, chen2023judgment}, ours is the first (to our knowledge) that proposes a direct uncertainty quantification approach.  \looseness=-1

\section{Conclusion}  

  We argued that indeterminacy is an inherent feature of some LLM evaluation tasks. We then showed that the standard gold label approach to evaluating LLMs can severely underestimate the true performance of an LLM when items are indeterminate, and proposed an alternative method that produces bounds on the true model performance. More broadly, treating ambiguity and vagueness as a meaningful source of signal opens the door to new paradigms for LLM evaluation. For instance, future work might develop tools to measure the \textit{uncertainty reduction} obtained by various improvements to the design of an evaluation---e.g., adding context to ambiguous items, refining definitions to reduce vagueness, or collecting additional (either human or AI) ratings. Such tools might help evaluation designers provide more robust assurances of performance and triage limited resources effectively.\looseness=-1
  \looseness=-1

\section*{Broader Impacts} 

A growing body of work reports ad hoc evaluations of LLMs. However, the ML community currently lacks a clear conceptual understanding of the LLM evaluation pipeline. Our DAG offers a step in this direction by distilling prior empirical work into key factors involved in the design of LLM evaluations. Although we focused on \textit{one} application of our framework (i.e., evaluating LLMs under task indeterminacy), our DAG also supports rigorous evaluation practices in a range of other areas---i.e., prompt robustness testing, prompt optimization, and annotation guideline refinement. Future work might operationalize our framework into a set of statistical evaluation tools that offer reliable, cost-effective LLM evaluations under the inherent ambiguity and vagueness present in many NLP tasks.\looseness=-1

\section*{Limitations}

One limitation of our framework (Figure \ref{fig:llm_eval}) is that its scope is limited to forced-choice NLP tasks. Although this is consistent with many safety and harm annotation workflows, it precludes more open-ended tasks (e.g., text summarization, open-ended QA tasks). Furthermore, while our framework parameterizes the effects of vagueness and ambiguity on LLM evaluations, it does not offer a comprehensive assessment of evaluation reliability and validity. For instance, a corpus can contain a collection of precisely-specified, unambiguous items while offering a flawed assessment of the capability or limitation being evaluated.\footnote{For example, consider an annotation guideline that instructs a rater to label a comment as \textit{``derogatory toward a person or a group of people’’} if it contains a keyword from a pre-defined dictionary of ``derogatory’’ terms. This task has a determinate response --- i.e., \textit{Yes} if the comment contains a keyword and \textit{No} otherwise. Yet, this approach may yield misleading evaluations if the dictionary of ``derogatory'' terms is poorly constructed.} Therefore, it is critical that our framework be used as part of a multifaceted evaluation protocol (i.e., including assessments of evaluation reliability and validity). \looseness=-1

Given the space constraints, we were unable to elaborate on all the components of our framework, nor were we able to provide worked case studies or framework applications illustrating its utility. Our framework was developed by synthesizing a diverse set of papers (N $>$ 80) on evaluating LLMs spanning the HCI (e.g., CHI, CSCW), ML (e.g., ICLR, NeurIPS, ICML), and NLP (e.g., ACL, EMNLP) communities.  We reviewed a subset of these papers in detail to understand their LLM evaluation setups and the assumptions of their methodologies. Nevertheless, our DAG may omit factors that are important in some evaluation contexts. Thus, our framework should not be viewed as exhaustive.\looseness=-1

{
\small
\bibliography{references}
\bibliographystyle{plainnat}
}

\end{document}